\documentclass[10pt]{amsart} 
\usepackage{amsmath,amssymb,bm, color} 
\usepackage{amsmath}
\usepackage{graphicx}
\usepackage{wrapfig}
\usepackage{graphicx} 
\usepackage{mathrsfs}
\usepackage{bbm}
\usepackage{bm}
\usepackage{amsfonts,amssymb}
\usepackage{multirow}
\usepackage{lineno}
\usepackage{color}

\numberwithin{equation}{section}

\usepackage{amsmath}
\usepackage{graphicx}
\usepackage[colorlinks=true, allcolors=blue]{hyperref}
\usepackage{amsfonts}
\usepackage{amssymb}
\usepackage{bm}
\usepackage{caption}
\usepackage{subfigure}
\usepackage{graphicx}
\usepackage{algorithm}
\usepackage{algorithmic}
\usepackage{multirow}
\usepackage{xcolor}
\usepackage{booktabs}
\usepackage{yfonts}
\usepackage{enumitem}
\usepackage{adjustbox}
\usepackage{mathtools}
\usepackage{float}
\usepackage{amsthm}
\usepackage{csquotes}
\newcommand{\Btheta}{\bm{\theta}}
\newcommand{\Bx}{\bm{x}}
\newcommand{\Be}{\bm{e}}

\newcommand{\Bp}{\bm{p}}

\title{Learning Epidemiological Dynamics via the Finite Expression Method}

\author {Jianda Du}
\address{Department of Mathematics, University of Florida, Gainesville, FL 32611, USA.}   
\email{jianda.du@ufl.edu}

\author {Senwei Liang}
\address{Lawrence Berkeley National Laboratory, 
Berkeley, CA 94720}   \email{SenweiLiang@lbl.gov} 

  \author {Chunmei Wang}
  \address{Department of Mathematics, University of Florida, Gainesville, FL 32611, USA. }
  \email{chunmei.wang@ufl.edu}
  \thanks{The research of Chunmei Wang was partially supported by National Science Foundation Grant DMS-2206332.} 
 
\begin{document}

\begin{abstract} 
Modeling and forecasting the spread of infectious diseases is essential for effective public health decision-making. Traditional epidemiological models rely on expert-defined frameworks to describe complex dynamics, while neural networks, despite their predictive power, often lack interpretability due to their ``black-box" nature. This paper introduces the Finite Expression Method, a symbolic learning framework that leverages reinforcement learning to derive explicit mathematical expressions for epidemiological dynamics. Through numerical experiments on both synthetic and real-world datasets, FEX demonstrates high accuracy in modeling and predicting disease spread, while uncovering explicit relationships among epidemiological variables. These results highlight FEX as a powerful tool for infectious disease modeling, combining interpretability with strong predictive performance to support practical applications in public health.
\medskip

\end{abstract}
 
\maketitle

\section{Introduction}
Partial differential equations (PDEs) and ordinary differential equations (ODEs) are used to describe physical phenomena across diverse scientific and engineering disciplines. In epidemiology, PDEs and ODEs serve as cornerstones for understanding disease dynamics, guiding interventions, and improving strategies to mitigate the impact of infectious diseases
\cite{anderson1991infectious, brauer2019mathematical}. Classical models such as the SIR (Susceptible-Infected-Recovered)~\cite{hethcote2000mathematics, kermack1927contribution}, SEIR (Susceptible-Exposed-Infected-Recovered)~\cite{anderson1991infectious, kermack1927contribution}, and SEIRD  (Susceptible-Exposed-Infected-Recovered- Deceased) \cite{brauer2019mathematical} frameworks rely on compartmental approaches, using differential equations to represent transitions between population states with parameters like transmission and recovery rates. These frameworks have underpinned epidemiological research for decades, offering valuable insights into disease spread and control.  

However, traditional compartmental models encounter significant limitations in addressing the complexities of real-world scenarios. Incorporating factors such as time-varying transmission rates, spatial heterogeneity, or additional compartments often makes these models analytically intractable and computationally intensive~\cite{keeling2008modeling}. Moreover, their reliance on manual refinements slows response times and hampers adaptability to rapidly evolving conditions~\cite{grassly2008mathematical, teobaldi2017proceedings}.

The emergence  of data-driven approaches, particularly those leveraging deep learning~\cite{huang2024attns,liang2022stiffness,liu2024training,ma2019model}, has introduced powerful alternatives for modeling epidemiological dynamics. Neural networks (NN)-based approaches, such as recurrent neural networks (RNNs)~\cite{hochreiter1997long}, have shown promise in capturing intricate patterns within epidemiological data ~\cite{goodfellow2016deep, lecun2015deep, raissi2019physics}. These methods enable rapid learning of disease dynamics, facilitating faster predictions and decision-making. However, their ``black-box" nature limits interpretability, hindering their utility in understanding the mechanisms driving disease spread~\cite{rudin2019stop}. Additionally, the implicit biases in NN optimization often favor smooth functions with rapid frequency decay, restricting their ability to produce highly accurate solutions~\cite{liang2024reproducing, CiCP-28-1746}.

Symbolic learning has recently emerged as a promising alternative, bridging the gap between the predictive power of machine learning and the interpretability of traditional models. By discovering governing equations directly from data, symbolic learning maintains mathematical rigor while leveraging data-driven insights ~\cite{cranmer2020discovering, schmidt2009distilling}.
The Finite Expression Method (FEX), introduced by Liang et al.~\cite{liang2022finite}, marks a significant advancement in symbolic learning for high-dimensional problems. FEX formulates the task of identifying mathematical expressions as a combinatorial optimization (CO) problem and leverages reinforcement learning (RL) to solve it. This approach automates the discovery of governing equations, drastically reducing development time while preserving interpretability and physical consistency. By producing parsimonious mathematical expressions, FEX becomes a powerful tool for reliable and efficient epidemiological modeling, particularly in addressing emerging public health crises.

This paper investigates the application of the FEX method for learning epidemiological models from both synthetic and real-world data. We demonstrate FEX's potential to address critical limitations of traditional and NN-based approaches. Specifically, our contributions include:

\begin{itemize}  
    \item \textbf{Advantages over NN-based methods:} We compare the FEX method with NN-based approaches, including RNNs, across three classical epidemiological models—SIR, SEIR, and SEIRD. FEX not only achieves competitive predictive performance but also derives explicit governing equations, offering superior interpretability. 

    \item \textbf{Advantages over traditional modeling approaches:} Using real-world COVID-19 data, we compare FEX with the fractional-order SEIQRDP model~\cite{bahloul2020fractional}. FEX demonstrates its versatility in handling complex epidemiological dynamics, providing actionable insights and enabling rapid model development.  
   
\end{itemize}

The remainder of this paper is organized as follows. Section~\ref{sec:overview} introduces the foundational concepts and procedural framework of the FEX method, supplemented by a detailed flowchart. Section~\ref{sec:loss} describes the loss function and optimization strategy used by FEX to derive governing equations for synthetic and real-world epidemiological data. Section~\ref{sec:result} presents experimental results, benchmarking FEX against both NN-based and traditional methods. Finally, Section~\ref{sec:conclusion} discusses the broader implications of our findings, acknowledges current limitations, and outlines directions for future research on extending FEX to other dynamical systems.

\section{The Finite Expression Methods}\label{sec:overview}
\begin{figure} 
\centering
\includegraphics[width=1\linewidth]{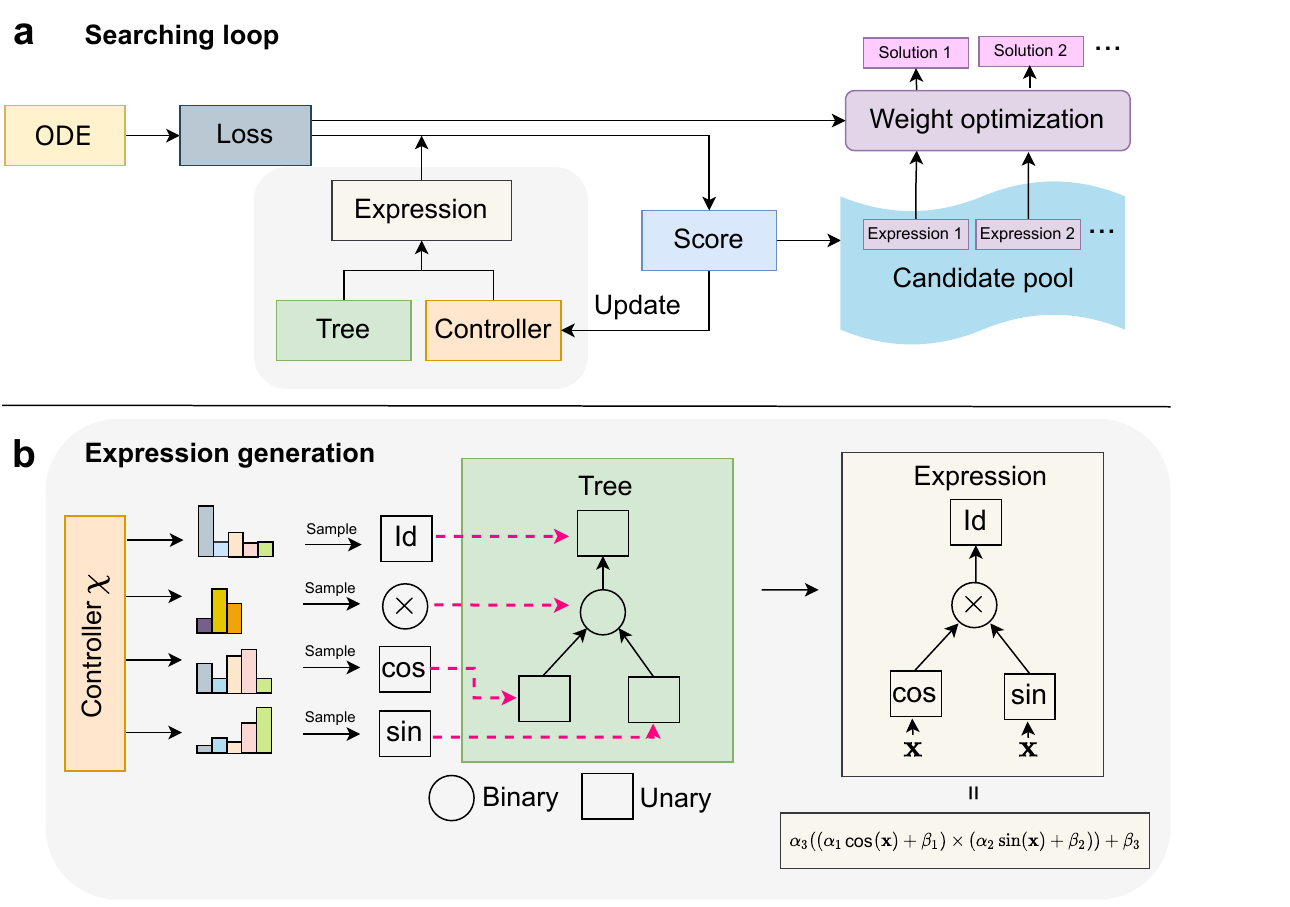}
    \caption{Flowchart of the FEX method. The process consists of an iterative search loop (a), weight optimization (b), and expression generation to identify solutions for the target ODEs or PDEs. Key components include the Expression Tree, Controller, and Candidate Pool, which collaboratively refine expressions through sampling, scoring, and optimization mechanisms.}
    \label{fig:symbo}
\end{figure}

The FEX method~\cite{jiang2023finite, liang2022finite,  song2024finite} provides a versatile framework for identifying governing equations of dynamical systems. It explores a function space composed of finite mathematical expressions constructed from a predefined set of operators. These expressions are represented as binary trees, denoted by $\mathcal{T}$ (Figure~\ref{fig:tree}), where each node corresponds to an operator, forming an operator sequence $\mathbf{e}$. Each operator is associated with trainable scaling and bias parameters, $\bm{\alpha}$ and  $\bm{\beta}$, collectively denoted as $\bm{\theta}$. This setup enables the representation of a finite expression as  $f(\bm{x}; \mathcal{T}, \mathbf{e}, \bm{\theta})$. The FEX framework seeks to identify governing equations by minimizing a functional $\mathcal{L}$ (e.g., derived from ODEs or PDEs). Formally, this optimization problem is expressed as:
$$\min \{\mathcal{L}(f(\cdot; \mathcal{T}, \Be, \bm{\theta}))|\Be, \bm{\theta}\}. $$
To solve this  CO problem, FEX employs a search loop powered by  RL, as illustrated in Figure~\ref{fig:symbo}a. The search process comprises four key components:

\begin{figure} 
\centering
\includegraphics[width=0.9\linewidth]{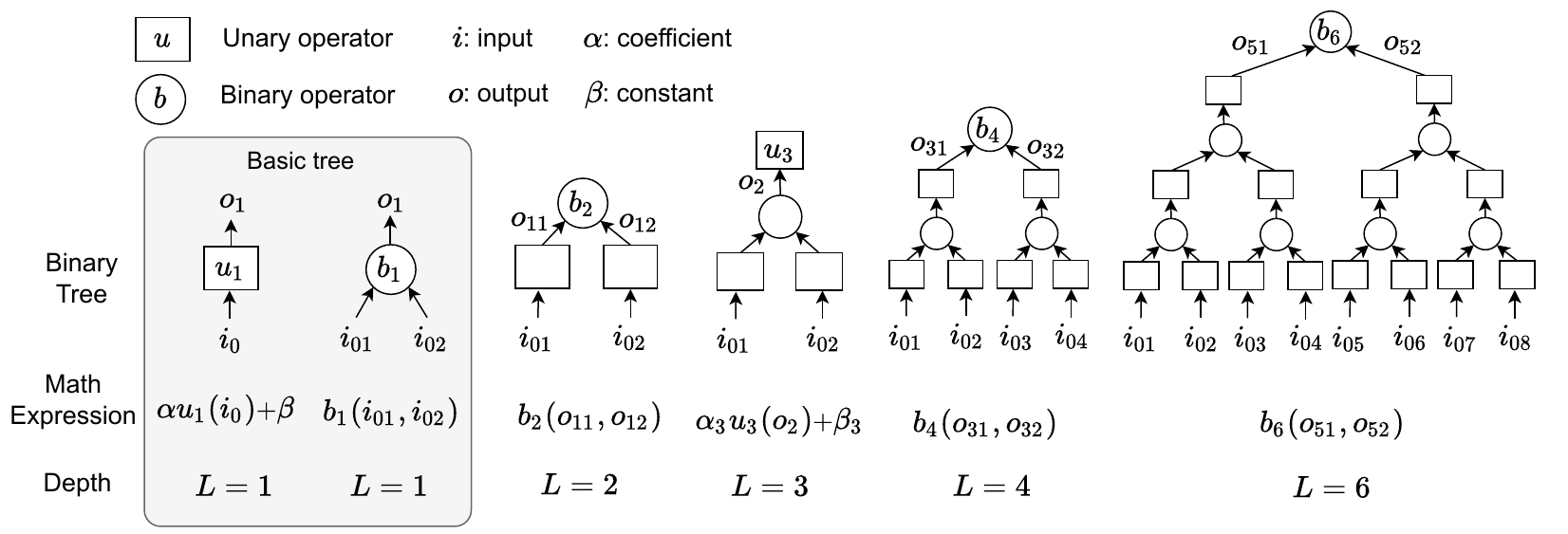}
\caption{The computational structure is represented using binary trees, where each node is assigned either a binary or unary operator. Expressions are recursively constructed, starting from depth-1 trees. Binary operators include  $\mathbb{B}:=\{+, -, \times, \div, \ldots\}$, and unary operators include  $\mathbb{U}:=\{\sin, \exp, \log, \text{Id}, (\cdot)^2, \int\cdot\text{d} x_i, \frac{\partial\cdot}{\partial x_i}, \ldots\}$.}
\label{fig:tree}
\end{figure}

  \textbf{Score Computation (Rewards in RL):}  
  The suitability of each operator sequence $\mathbf{e}$ is assessed via a score, $S(\Be)$, which is defined as: 
  $$      S(\Be) := \big(1+L(\Be)\big)^{-1},$$ 
    where $L(\Be) := \min \{\mathcal{L}(f(\cdot; \mathcal{T}, \Be, \bm{\theta}))|\bm{\theta}\}$. A smaller  $L(\Be)$ indicates better approximation of the governing dynamics, corresponding to a higher score.

   To address the computational challenges of minimizing $\mathcal{L}$ globally, FEX employs a two-stage optimization strategy. First, a first-order algorithm runs for $T_1$ iterations to provide an initial parameter estimate. Subsequently, a second-order algorithm, such as Broyden-Fletcher-Goldfarb-Shanno (BFGS)~\cite{fletcher2000practical}, refines these parameters over $T_2$ iterations. If  $\bm{\theta}_0^{\Be}$ represents the initial parameters and  $\bm{\theta}_{T_1+T_2}^{\Be}$ the refined ones, the score is approximated as:
     $$   S(\Be) \approx \big(1+\mathcal{L} (f(\cdot; \mathcal{T}, \Be, \bm{\theta}_{T_1+T_2}^{\Be}))\big)^{-1}. $$

  \textbf{Operator Sequence Generation (Actions in RL):}  
  The controller, denoted by $\bm{\chi}_\Phi$ generates operator sequences $\Be$ (Figure~\ref{fig:symbo}b). The parameters $\Phi$  of the controller are updated iteratively to favor high-scoring sequences. Operator sequences are constructed by sampling from probability mass functions $\Bp_\Phi^1, \Bp_\Phi^2, \ldots, \Bp_\Phi^s$, which define the distributions of node values in $\mathcal{T}$. A sequence $\Be = (e_1, e_2, \ldots, e_s)$ is generated by sampling each  $e_j$   from $\Bp_\Phi^j$,

  To encourage exploration, an $\epsilon$-greedy strategy~\cite{sutton2018reinforcement} is employed: with probability $\epsilon$, $e_i$ is sampled uniformly from the operator set, while with probability $1-\epsilon$, it is sampled from $\Bp_\Phi^i$. Larger values of $\epsilon$ promote broader exploration of the search space.

    \textbf{Controller Update (Policy Optimization in RL):}  
   The controller's parameters  $\Phi$ are updated to increase the likelihood of generating high-performing sequences. Using a policy gradient approach~\cite{petersen2019deep, ruder2016overview}, the objective is to maximize:
     $$   \mathcal{J}(\Phi)=\mathbb{E}_{\Be \sim \bm{\chi}_\Phi} \{S(\Be)|S(\Be)\geq S_{\nu, \Phi}\},$$
    where $S_{\nu, \Phi}$ is the $(1-\nu)\times 100\%$-quantile of the score distribution produced by $\bm{\chi}_\Phi$  in a batch. This prioritizes top-performing sequences. Gradient ascent is used to update 
      $$  \Phi \leftarrow \Phi+\eta \nabla_\Phi\mathcal{J}(\Phi).$$ 

  \textbf{Candidate Optimization (Policy Deployment):}  
    A candidate pool $\mathbb{P}$, with a fixed capacity $K$, stores the highest-scoring operator sequences discovered during the search. Additional optimization is performed for each sequence $\Be \in \mathbb{P}$, using a first-order algorithm over $T_3$ iterations with a smaller learning rate. This reduces the risk of missing promising solutions due to local minima encountered in the initial search.

\section{Learning Epidemiological Dynamics  via FEX}\label{sec:loss}
 
\subsection{Epidemiological Models} The dynamics of an epidemiological system are governed by: 
   $$ \frac{d\Bx}{dt} = f(\Bx), $$
where \(\Bx\in \mathbb{R}^d\) denotes the   state variables (e.g., susceptible, infected, and recovered populations), and 
$f(\Bx)$ encapsulates the system's temporal evolution.

To evaluate the effectiveness of the FEX method, we apply it to three commonly used epidemiological models:
 
 \textbf{(1) SIR Model:} The SIR model~\cite{hethcote2000mathematics, kermack1927contribution} divides the total population \(N\) into three compartments: susceptible (\(S\)), infectious (\(I\)), and recovered (\(R\)).  This model is suitable for diseases with long-term immunity post-recovery, such as measles or chickenpox. The governing equations are:
    \begin{equation}\label{sir}
\begin{split} 
    \frac{dS}{dt} =& \mu (N - S) - \frac{\beta S I}{N}, \\
    \frac{dI}{dt} =& \frac{\beta S I}{N} - (\mu + \gamma) I, \\
    \frac{dR}{dt} = &\gamma I - \mu R,
\end{split} 
\end{equation}
where  \(\beta\), \(\gamma\) and \(\mu\) are the transmission, recovery, and natural death rates, respectively.

  \textbf{(2) SEIR Model:} The SEIR model~\cite{anderson1991infectious, kermack1927contribution} extends the SIR framework by introducing an exposed (\(E\)) compartment to account for a latent, non-infectious period. Additionally, vaccination effects are incorporated through a $\nu S$ term. The dynamics are:
    \begin{equation}\label{seir}
\begin{split} 
    \frac{dS}{dt} =& \mu (N - S) - \frac{\beta S I}{N} - \nu S, \\
    \frac{dE}{dt} = &\frac{\beta S I}{N} - (\mu + \sigma) E, \\
    \frac{dI}{dt} = &\sigma E - (\mu + \gamma) I, \\
    \frac{dR}{dt} =& \gamma I - \mu R + \nu S, 
\end{split} 
\end{equation}
where \(\sigma\) is the rate of progression from exposed to infectious, and \(\nu\) is the rate of immunity acquisition.

  \textbf{(3) SEIRD Model:} The SEIRD model~\cite{brauer2019mathematical} further extends the SEIR by including a deceased (\(D\)) compartment to account for disease-induced mortality. The equations are:
    \begin{equation}\label{seird}
\begin{split} 
    \frac{dS}{dt} =& -\frac{\beta S I}{N},\\
    \frac{dE}{dt} =& \frac{\beta S I}{N} - \sigma E, \\
    \frac{dI}{dt} = &\sigma E - (\gamma + \delta) I, \\
    \frac{dR}{dt} =& \gamma I,\\
    \frac{dD}{dt} = &\delta I,
\end{split} 
\end{equation}
where $\delta$  is the disease-induced death rate.     This model is crucial for analyzing diseases with significant mortality.

In all models, the variables are normalized such that their sum equals 1 (\(N=1\)).

\subsection{Learning Epidemiological Dynamics via FEX}
Given historical data \( \{\Bx^{s\Delta}\}_{s=0}^M \), where $s$ is the step index, $M$ is the total number of steps, and $\Delta$ is the time step size, our goal is to construct a surrogate model \( \phi(\Bx): \mathbb{R}^d \to \mathbb{R}^d \) that approximates the true dynamics \(f\). 
This surrogate model \(\phi(\Bx)\) aims to  replicate the system's  behavior and predict future states. Let $\hat{\Bx}^{(s+1)\Delta} = \text{Integrator}(\phi, \Bx^{s\Delta}, \Delta)$  denote the predicted state, where the integrator advances the system by one time step $\Delta$, starting from $\Bx^{s\Delta}$. Accurate replication requires  $\hat{\Bx}^{(s+1)\Delta}\approx \Bx^{(s+1)\Delta}$, for $s=0,\cdots, M-1$. 

\textbf{Training:} The surrogate model is trained by minimizing the discrepancy between observed data and model predictions. The objective function is  
    $$\min_{\phi}\mathcal{L}(\phi):=\frac{1}{M} \sum_{s=0}^{M-1}  \| {\Bx}^{(s+1)\Delta} - \hat{\Bx}^{(s+1)\Delta} \|_2^2, $$
where $\phi$ belongs to a specified function space. For neural networks, $\phi_{\rm{NN}}(\Bx;\Btheta)$ is  parameterized by   $\Btheta$, and the objective becomes: 
\begin{align}\min_{\theta}\mathcal{L}(\phi_{\rm{NN}}(\Bx;\Btheta)).
\label{eqn:nnloss}\end{align}
This optimization can be performed using methods such as Adam or BFGS.

\textbf{Forecasting:} After training, the optimized surrogate model $\tilde{\phi}$  predicts future states beyond the training period by iteratively propagating the system forward. Starting from $\hat{\Bx}^{M\Delta}=\Bx^{M\Delta}$, future states are computed as: 
$$\hat{\Bx}^{(s+1)\Delta} = \text{Integrator}(\tilde{\phi}, \hat{\Bx}^{s\Delta}, \Delta), s\geq M.  $$

In epidemiological applications, the trained surrogate model can forecast disease spread and evaluate intervention strategies, such as vaccination or social distancing.

\textbf{FEX Methodology:}
FEX aims to approximate $f(\Bx)$ with a surrogate function:
    $$ \Phi_{\rm FEX}(\Bx):=[\phi^{(1)}_{\rm FEX}(\Bx), \phi^{(2)}_{\rm FEX}(\Bx),\cdots, \phi^{(d)}_{\rm FEX}(\Bx)]^\top\approx f(\Bx),$$
where $\phi^{(i)}_{\rm FEX}(\Bx):\mathbb{R}^d\to \mathbb{R}^1$ for $i=1,\cdots,d$.

Given historical data \( \{\Bx^{s\Delta}\}_{s=0}^M \), with $\Bx^{s\Delta}=[x^{s\Delta}_1, \cdots, x^{s\Delta}_d]$, the loss function for each component is:
 $$   \min_{\bm{e}, \bm{\theta}}\mathcal{L}(\phi^{(i)}_{\rm FEX}(\Bx; \bm{e}, \bm{\theta})):=\frac{1}{M} \sum_{s=0}^{M-1}  \| x_i^{(s+1)\Delta} - \hat{x}_i^{(s+1)\Delta}  \|_2^2, $$
where $\hat{x}_i^{(s+1)\Delta}$ is predicted with the surrogate model starting from $\Bx^{(s+1)\Delta} $. 

Using an Euler scheme, the loss function for $\phi^{(i)}_{\rm FEX}(\Bx)$ becomes:
\begin{equation}
    \label{eqn:mseloss}
    \mathcal{L}(\phi^{(i)}_{\rm FEX}(\Bx; \bm{e}, \bm{\theta})):=\frac{1}{M} \sum_{s=0}^{M-1}  \| x_i^{(s+1)\Delta} - x_i^{s\Delta}- \phi^{(i)}_{\rm FEX}(\Bx^{s\Delta}) \Delta  \|_2^2. 
\end{equation}
This loss function allows independent learning of each component of the dynamics using the RL-based optimization approach detailed in  Section~\ref{sec:overview}.

\section {Numerical results}\label{sec:result}

In this section, we demonstrate the effectiveness of the FEX method in learning epidemiological dynamics using both synthetic and real-world data.

\subsection{Synthetic Epidemiological Data}  Synthetic data is generated based on the SIR~\eqref{sir}, SEIR~\eqref{seir}, and SEIRD~\eqref{seird} models. The performance of the FEX method is evaluated against neural network-based approaches, specifically the NN and RNN methods~\cite{goodfellow2016deep, lecun2015deep, sutton2018reinforcement}. The NN method uses a neural network as a surrogate to approximate the dynamics, while the RNN method leverages a recurrent neural network for modeling time-series data. Model performance is quantified using the mean squared error (MSE) at each time step, calculated between the predicted and true trajectories generated from varying initial conditions.

\textbf{Data Generation:} 
The parameter values in the SIR model \eqref{sir}, the SEIR model \eqref{seir}, and the SEIRD model \eqref{seird} are set as:  \(\beta = 0.9\), \(\gamma = 0.2\), \(\mu = 0.3\), \(\sigma_1 = 0.6\)(for SEIR), \(\sigma_2 = 0.5\)(for SEIRD), \(\nu = 0.2\) and \(\delta = 0.05\).  Data for the three models is generated using Euler's method. For each model, 200 simulated trajectories are generated, each containing $M=250$ time steps with a time step size of \(\Delta = 0.2\). These trajectories are divided evenly into training and testing datasets. Initial conditions for each trajectory are sampled from a uniform distribution  $\mathcal{U}(0, 1)$.  

\textbf{Training  Procedures:}\label{sec:trees}

\begin{figure}
     \centering    \includegraphics[width=0.7\linewidth]{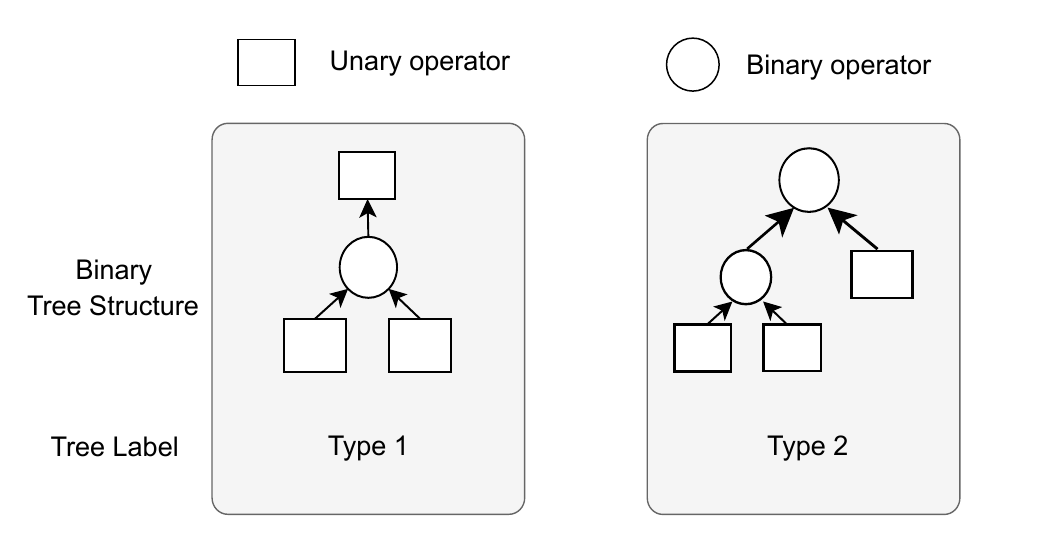}
     \caption{Illustration of two tree structures  used in the FEX implementation. }
     \label{fig:Two_trees}
 \end{figure}

\textbf{The FEX method:} Two types of tree structures are considered (see Figure \ref{fig:Two_trees}).
Type 1 consists of three layers with one binary operator and three unary operators, while Type 2 also has three layers but includes two binary operators and three unary operators. Type 2 is used to approximate differential equations with nonlinear terms, while Type 1 is employed to  approximate differential equations without nonlinear terms aiming to achieve the desired form with fewer training iterations. The FEX model is trained for 100 epochs with a batch size of 10. A greedy search strategy~\cite{feo1995greedy, liang2022finite, wilt2010comparison} with a probability of 0.1 is adopted, and the learning rate for optimizing the controller is set to 0.002. The candidate unary operators include \( 0, 1, x, x^2, x^3, x^4, \sin(x), \cos(x), \exp(x) \), while the candidate binary operators are \( +, -, \times \). Euler's method is applied as the integrator in the loss function \eqref{eqn:mseloss}.

\textbf{The NN method:} For the NN method as described in~\eqref{eqn:nnloss}, a network with three linear layers and two ReLU activation layers is employed. The NN model is trained using the Adam optimizer with a learning rate of 0.001 for 100 epochs and a batch size of 32. Euler's method~\cite{burden1997numerical, shen2020deep} is used as the integrator in the loss function \eqref{eqn:mseloss}.

\textbf{The RNN method:} The RNN model consists of two Long Short-Term Memory (LSTM) layers and a linear output layer for predicting time-series data of the variables in the epidemiological models. The input to the RNN model comprises the variables from the training dataset, where the input dimension corresponds to the number of variables in the specific epidemiology model (e.g., 3 for SIR, 4 for SEIR, and 5 for SEIRD). Each LSTM layer has a hidden size of 51, and the final linear layer maps the 51-dimensional hidden state to an output dimension matching the number of variables in the corresponding model. At the beginning of each forward pass, the hidden and cell states of both LSTM layers are initialized to zero tensors. The model is trained using the limited-memory BFGS (LBFGS) optimizer with the MSE loss function.

\textbf{Testing:} After identifying the expressions that best fit the training data, predicted trajectories are generated by applying Euler's method to the initial state of each testing trajectory. The testing MSE is calculated by comparing the predicted trajectories with the actual testing data.

\begin{figure} 
\centering
\subfigure[SIR]{\label{fig:sir}\includegraphics[width=0.45\linewidth]{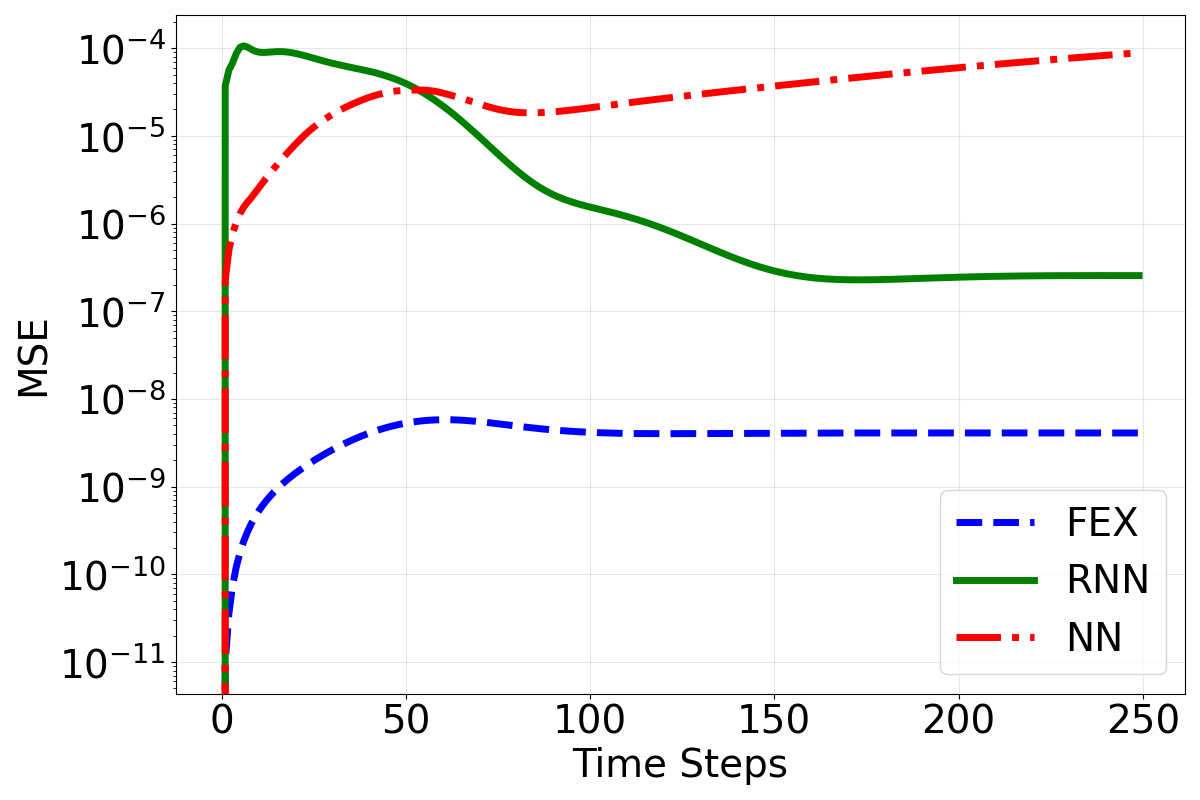}}
\subfigure[SEIR]{\label{fig:seir}\includegraphics[width=0.45\linewidth]{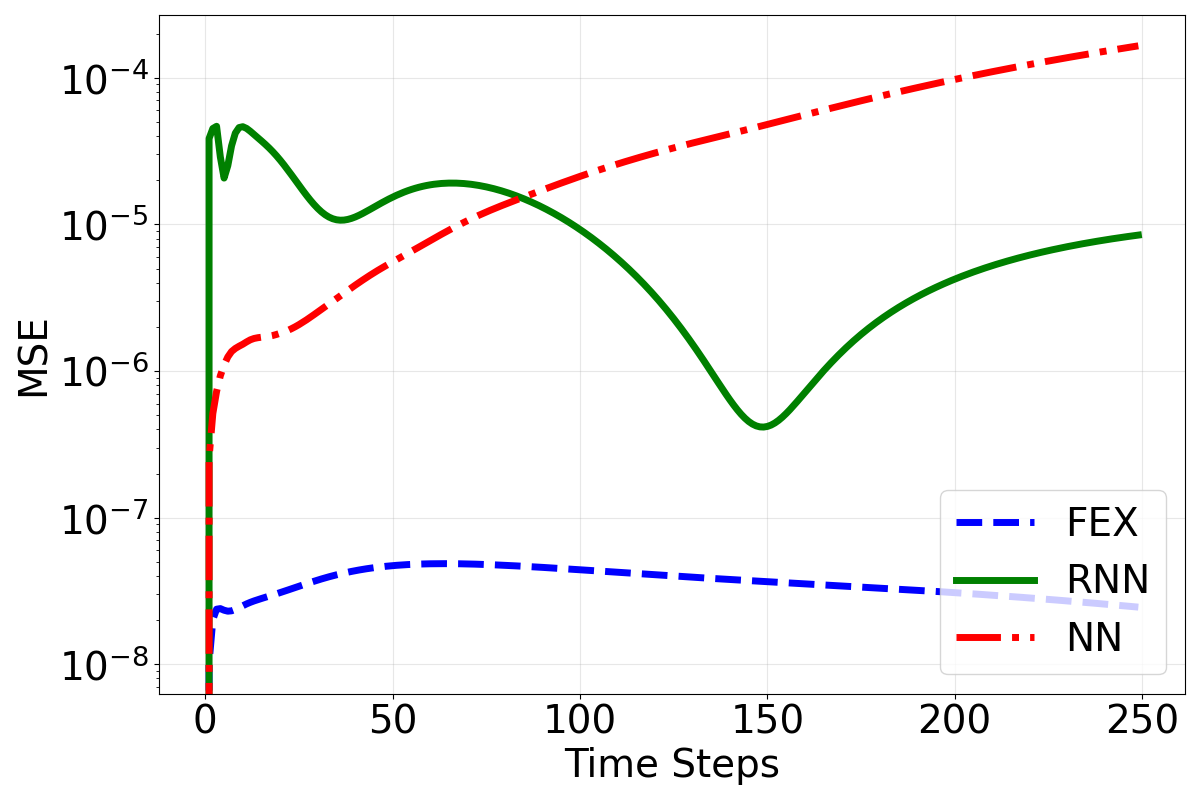}}
\subfigure[SEIRD]{\label{fig:seird}\includegraphics[width=0.45\linewidth]{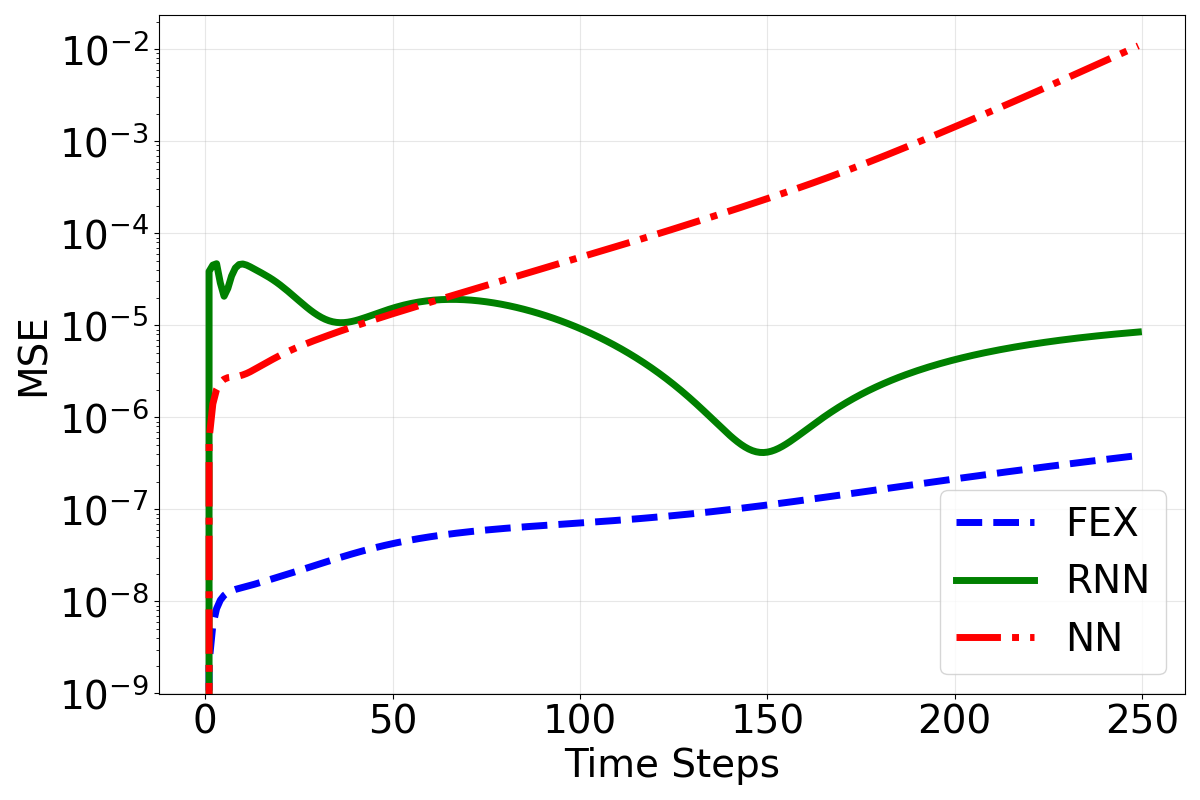}}
\caption{Comparison of MSE over time for three methods (FEX, RNN, NN) on   (a) SIR model, (b) SEIR model, and (c) SEIRD model  over 250 time steps.}
\label{fig:4}
\end{figure}

\textbf{Numerical Results:}  
Figure \ref{fig:4}  presents the testing MSE over time, comparing the performance of FEX with the NN and RNN methods on the SIR, SEIR, and SEIRD models, respectively. The results show that the FEX method achieves the smallest MSE on the order of \(10^{-8}\sim 10^{-7}\), significantly outperforming the NN and RNN models, which exhibit MSEs on the order of \(10^{-4}\sim 10^{-2}\) and \(10^{-6}\sim 10^{-5}\), respectively. Additionally, the FEX method maintains consistently small errors over time, whereas the error trajectories predicted by the NN method increase as time progresses. These findings highlight the superior predictive accuracy and robustness of the FEX method.

\subsection{Real-World Epidemiological Data}
To validate the FEX method on real-world data, we use the publicly available COVID-19 dataset from Our World in Data~\cite{jhu_covid19_dataset}, which aggregates detailed global case records from the Johns Hopkins University COVID-19 Data Repository. We compare the FEX method to the Fractional-Order SEIQRDP model~\cite{bahloul2020fractional}, a well-regarded approach for modeling COVID-19 dynamics using fractional-order derivatives to capture complex epidemic patterns.

COVID-19, caused by the highly transmissible SARS-CoV-2 virus, emerged in late 2019 and rapidly escalated into a global pandemic. Understanding its transmission dynamics, influenced by interactions among susceptible, exposed, infected, and recovered individuals, is essential for devising public health interventions. By applying FEX to real-world COVID-19 data, we demonstrate its effectiveness and interpretability compared to established methods.

\textbf{Data Acquisition:} We analyze daily COVID-19 data from Hubei, China, focusing on active cases ($Q$), deceased cases ($D$), and recovered cases ($R$). The dataset spans 1,147 days and is publicly accessible in~\cite{bahloul2020fractional}. For this study, we use data from the first 100 days (January 22 to April 30, 2020), as values stabilize beyond this period. The dataset is split into training (first 85 days) and testing (remaining 15 days) subsets.

\textbf{Training:} Using the FEX framework, we model COVID-19 dynamics with a Type 2 tree structure (Figure~\ref{fig:Two_trees}). Training is conducted over 100 epochs with a batch size of 10, employing a greedy search strategy with a 0.1 exploration probability and a controller learning rate of 0.002. For comparison, the fractional-order SEIQRDP method is implemented using the MATLAB code provided in~\cite{bahloul2020fractional}. Both methods are trained on the first 85 days of data and evaluated on the subsequent 15 days.

\textbf{Testing:} 
To assess the FEX model’s performance, we use the one-step Euler method~\cite{burden1997numerical} to simulate both training and prediction trajectories:

\begin{itemize}
    \item Training Trajectory ($t=0$ to
$t=84$): Starting with the ground truth at $t = 0$, the model computes subsequent values using the one-step Euler method and ground truth inputs from the previous time step.

\item Prediction Trajectory ($t=85$ to
$t=99$): At $t=85$, predictions are initiated using the one-step Euler method and the ground truth from $t = 84$. For later steps, predictions are generated iteratively using outputs from the previous time step as inputs.

\end{itemize}
This approach evaluates the FEX method’s ability to generalize beyond the training phase to unseen data.

\begin{figure} 
\centering
\subfigure[Active]{\label{fig:A}\includegraphics[width=0.47\linewidth ]{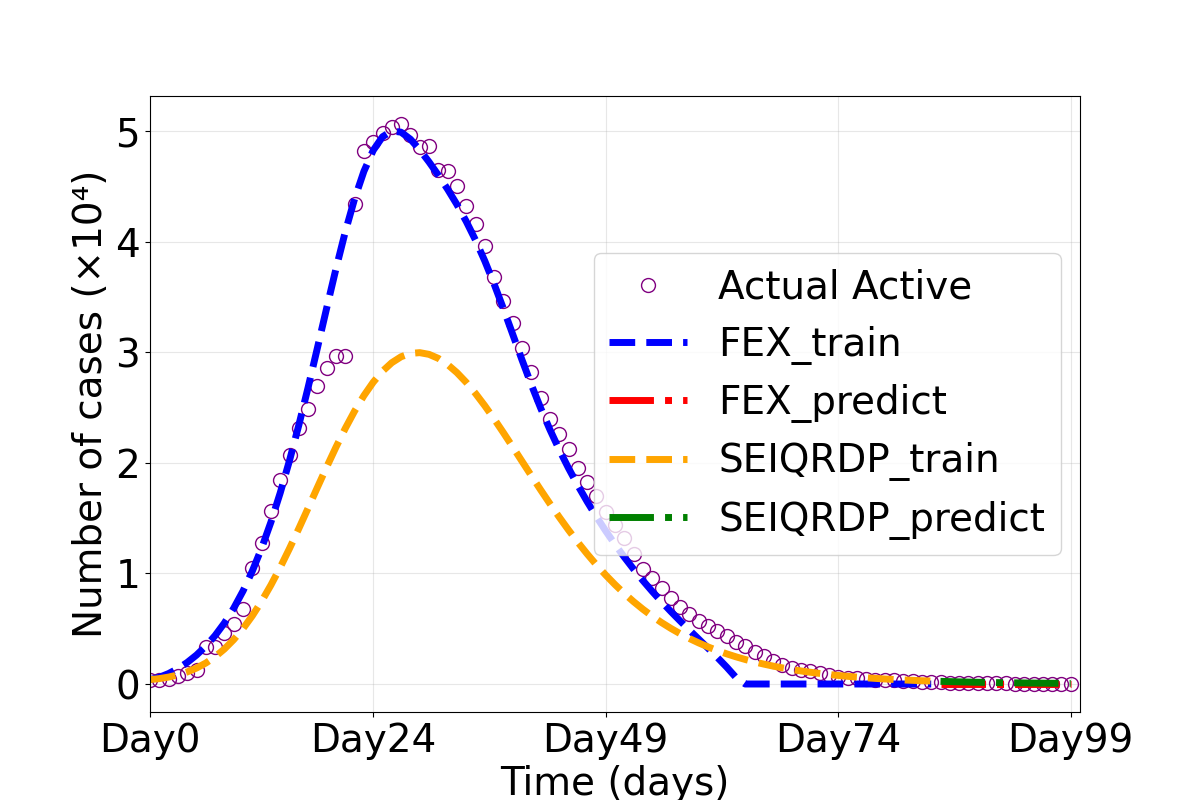}}
\subfigure[Deceased]{\label{fig:D}\includegraphics[width=0.47\linewidth ]{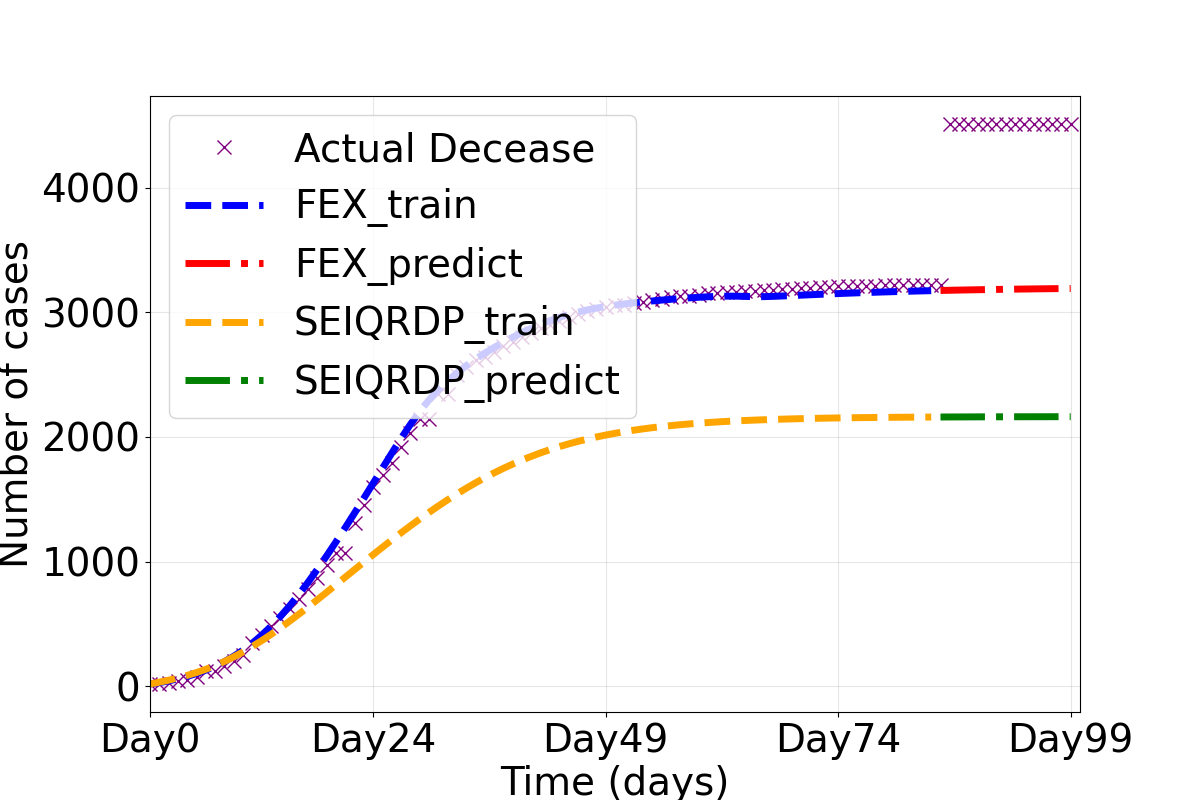}}
\subfigure[Recovered]{\label{fig:R}\includegraphics[width=0.47\linewidth ]{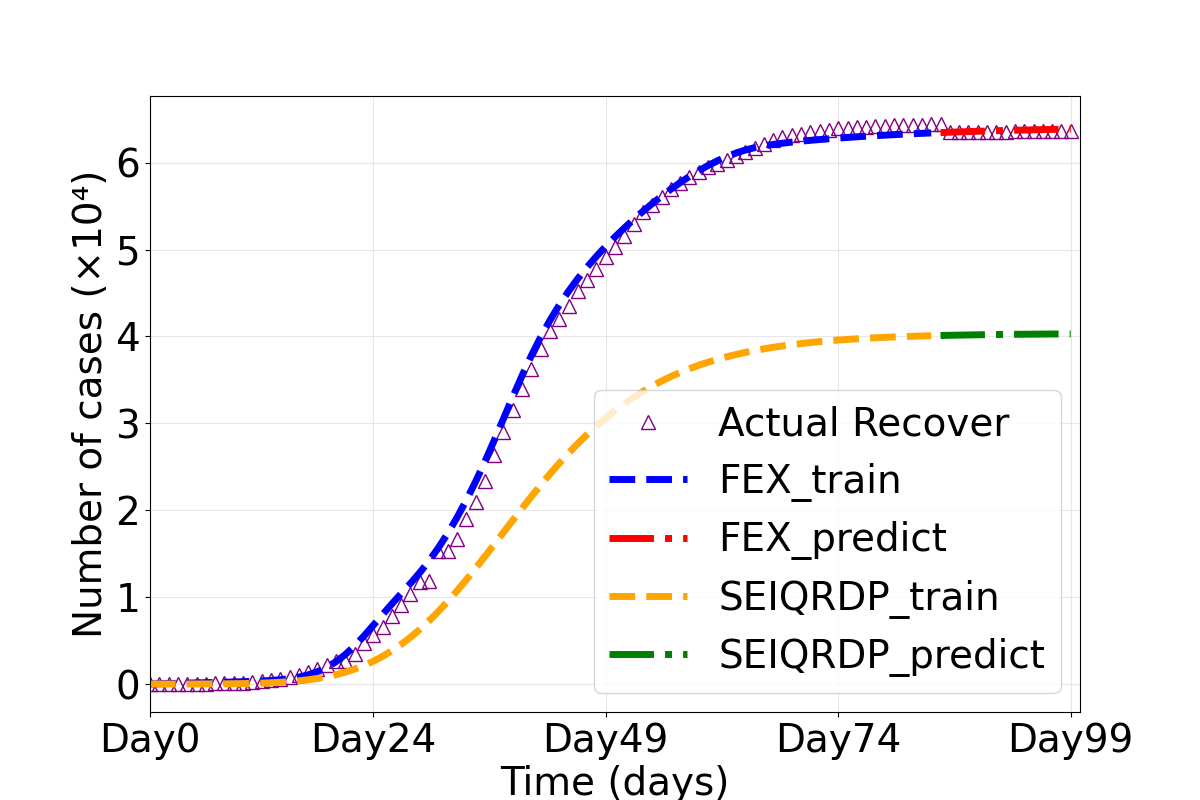}}
\caption{Comparison of actual and predicted COVID-19 cases using the FEX and SEIQRDP  methods: (a) active cases ($Q$), (b) deceased cases ($D$), and (c) recovered cases ($R$). FEX demonstrates superior data fitting during the training phase and accurate predictions in the testing phase, compared to SEIQRDP.}
\label{fig:5}
\end{figure}

\textbf{Numerical Results:}
The FEX method identifies the governing dynamics of $R$, $D$, and $Q$ during the training phase, effectively capturing interactions among these critical epidemiological variables. The learned equations are:
\begin{equation*}
\begin{split} 
    \frac{dR}{dt} &= 
    \big(-0.9030 R^3 + 2.4025 D^3 - 0.0262 Q^3 + 0.0311 \big) \nonumber \\
    &\quad \cdot \big(-0.1840 R^3 - 0.0432 D^3 - 2.5147 Q^3 - 0.0181 \big) \nonumber \\
    &\quad \cdot \Big(0.1919 \sin(R) + 0.1812 \sin(D) + 0.7006 \sin(Q) - 0.7283\Big), 
\end{split}
\end{equation*}
\begin{equation*}
\begin{split}
    \frac{dD}{dt} &= 
    (-1.5383 e^R + 1.3790 e^D - 0.0797 e^Q - 0.9186)  \\
    &\quad \cdot (-0.4292 R + 0.9682 D - 0.2423 Q - 0.2602) \\
    &\quad \cdot (-0.8972 R + 1.4067 D - 0.2637 Q + 0.0054), 
\end{split}
\end{equation*}
\begin{equation*}
\begin{split} 
    \frac{dQ}{dt} &= 
    \Big(1.6211 \sin(R) - 2.1673 \sin(D) + 0.6259 \sin(Q) - 0.0008 \Big)   \\
    &\quad \cdot \big(4.3940 R^3 + 1.6576 D^3 + 0.5903 Q^3 - 0.6928\big) \nonumber \\
    &\quad \cdot \big(0.1196 R^2 - 3.7194 D^2 + 1.7070 Q^2 + 1.7746\big). 
    \end{split}
\end{equation*}
Figure  \ref{fig:5}  compares FEX and the SEIQRDP method~\cite{bahloul2020fractional} on $Q$, $D$, and $R$. During training, FEX achieves superior data fitting. In the prediction phase, FEX closely aligns with observed data, demonstrating robust generalization. Furthermore, FEX uses only three input variables ($Q$, $R$, and $D$) to derive governing equations, while the SEIQRDP method requires five ($E$, $I$, $Q$, $R$, and $D$) in general, with assumptions $E = I$ and $I = Q+R+D$ for current data. This highlights FEX’s simplicity and efficiency in modeling complex epidemiological dynamics.

\section{Conclusions and Future Work}\label{sec:conclusion}

This paper introduces the Finite Expression  Method for learning epidemiological dynamics directly from data. The effectiveness of FEX is demonstrated through extensive experiments on both synthetic datasets—generated from SIR, SEIR, and SEIRD models—and real-world COVID-19 data. The results show that FEX not only achieves high accuracy in modeling dynamics but also produces interpretable mathematical expressions that approximate the underlying relationships. These expressions provide valuable insights into the interactions among different population groups in epidemiological data, enabling a deeper understanding of disease spread.

Despite its promising results, FEX faces several challenges when applied to learning epidemiological dynamics:
\begin{itemize}
    \item \textbf{Computational Cost:} The primary computational burden stems from the search process. Specifically, evaluating the score of each candidate operator sequence requires performing optimization from scratch, which is computationally intensive. This complexity can be mitigated using parallelization techniques to accelerate the search process.

    \item 
 \textbf{Nonuniqueness of Solutions:} FEX may generate multiple valid finite expressions for the same dataset. While these solutions can still be meaningful, the lack of uniqueness complicates interpretability and may pose challenges for drawing definitive conclusions. Addressing this issue could involve incorporating improved regularization techniques and increasing the size and diversity of training data to enhance the consistency and reliability of the generated solutions.
\end{itemize}  

 \textbf{Future work} will focus on addressing these challenges, including the development of more efficient search algorithms and techniques to promote solution uniqueness. These advancements aim to enhance the utility and robustness of the FEX method in practical epidemiological applications.

\end{document}